\documentclass[10pt,twocolumn,letterpaper]{article}

\usepackage{cvpr}              

\makeatletter
\@namedef{ver@everyshi.sty}{}
\makeatother

\usepackage[dvipsnames,svgnames,x11names]{xcolor}
\usepackage{tikz}
\usetikzlibrary{arrows.meta,shapes,calc,matrix,fit,positioning,backgrounds,decorations.markings}
\usepackage{pgfplots}
\usepackage{pgfplotstable}
\pgfplotsset{compat=1.9}
\usepackage{xstring}

\usepgfplotslibrary{external}

\IfBeginWith*{\jobname}{fig/extern/}{\finalcopy}{}

\tikzstyle{every picture}+=[
	remember picture,
	every text node part/.style={align=center},
	every matrix/.append style={ampersand replacement=\&},
]
\tikzstyle{tight} = [inner sep=0pt,outer sep=0pt]
\tikzstyle{node}  = [draw,circle,tight,minimum size=12pt,anchor=center]
\tikzstyle{op}    = [draw,circle,tight]
\tikzstyle{dot}   = [fill,draw,circle,inner sep=1pt,outer sep=0]
\tikzstyle{pt}    = [fill,draw,circle,inner sep=1.5pt,outer sep=.2pt]
\tikzstyle{box}   = [draw,rectangle,inner sep=3pt]
\tikzstyle{high}  = [black!60]
\tikzstyle{group} = [high,box,opacity=.5]
\tikzstyle{dim1}  = [fill opacity=.3,text opacity=1]
\tikzstyle{dim2}  = [fill opacity=.5,text opacity=1]
\tikzstyle{dim3}  = [fill opacity=.7,text opacity=1]
\tikzstyle{rectc} = [tight,transform shape]
\tikzstyle{rect}  = [rectc,anchor=south west]

\tikzset{every mark/.append style={solid}}
\pgfplotsset{
	grid=both, width=\columnwidth, try min ticks=5,
	every axis/.append style={font=\small},
	every axis plot/.append style={thick,mark=none,mark size=1.8,tension=0.18},
	legend cell align=left, legend style={fill opacity=0.8},
	xticklabel={\pgfmathprintnumber[assume math mode=true]{\tick}},
	yticklabel={\pgfmathprintnumber[assume math mode=true]{\tick}},
	nodes near coords math/.style={
		nodes near coords={\pgfmathprintnumber[assume math mode=true]{\pgfplotspointmeta}},
	},
}

\pgfplotsset{
	dash/.style={mark=o,dashed,opacity=0.6},
	dott/.style={mark=o,dotted,opacity=0.6},
	nolim/.style={enlargelimits=false},
	plain/.style={every axis plot/.append style={},nolim,grid=none},
}

\pgfdeclarelayer{bg4}
\pgfdeclarelayer{bg3}
\pgfdeclarelayer{bg2}
\pgfdeclarelayer{bg1}
\pgfdeclarelayer{fg1}
\pgfdeclarelayer{fg2}
\pgfdeclarelayer{fg3}
\pgfdeclarelayer{fg4}
\pgfsetlayers{bg4,bg3,bg2,bg1,main,fg1,fg2,fg3,fg4}

\pgfkeys{/tikz/.cd, aspect/.store in=\aspect, aspect=1}
\pgfkeys{/tikz/.cd, depth/.store in=\depth, depth=.5}
\pgfkeys{/tikz/.cd, stepx/.store in=\step, stepx=1}

\tikzstyle{geom} = [line join=bevel,aspect=1,depth=.5,z={(\depth*\aspect,\depth)}]
\tikzstyle{wire} = [geom,draw,thick]

\def\cx[#1,#2,#3]{#1}
\def\cy[#1,#2,#3]{#2}
\def\cz[#1,#2,#3]{#3}
\def\ex[#1,#2,#3]{#1,0,0}
\def\ey[#1,#2,#3]{0,#2,0}
\def\ez[#1,#2,#3]{0,0,#3}

\usepackage{times}
\usepackage{epsfig}
\usepackage{graphicx}
\usepackage{amsmath}
\usepackage{amssymb}
\usepackage{booktabs}
\usepackage{multirow}

\usepackage{comment}
\usepackage{color}
\usepackage[accsupp]{axessibility}

\usepackage{mathtools}
\usepackage{bm}

\usepackage{enumitem}
\usepackage{xspace}
\usepackage{breakcites}
\usepackage{contour}
\usepackage{pifont}
\usepackage{mathrsfs}

\usepackage[utf8]{inputenc}
\usepackage{kotex}
\usepackage{kotex-logo}

\usepackage{algorithm}
\usepackage{algpseudocode}
\usepackage{mycommands}

\usepackage[pagebackref=true,breaklinks=true,colorlinks,bookmarks=false]{hyperref}

\usepackage[capitalize]{cleveref}
\crefname{section}{Sec.}{Secs.}
\Crefname{section}{Section}{Sections}
\Crefname{table}{Table}{Tables}
\crefname{table}{Tab.}{Tabs.}

\def\cvprPaperID{9480} 
\def\confName{CVPR}
\def\confYear{2023}

\newcommand*\samethanks[1][\value{footnote}]{\footnotemark[#1]}

\begin{document}


\title{SyncMask: Synchronized Attentional Masking for Fashion-centric Vision-Language Pretraining}

\author{
Chull Hwan Song$^1$\thanks{Equal contribution} \ \ \ \ Taebaek Hwang$^1$\samethanks \ \ \ \ Jooyoung Yoon$^1$ \ \ \ \ Shunghyun Choi$^1$ \ \ \ \ Yeong Hyeon Gu$^{2}$\thanks{Corresponding author}\\
\normalsize $^1$Dealicious Inc.\quad
\normalsize $^2$Sejong University \\
}

\maketitle

\thispagestyle{empty}

\newcommand{\head}[1]{{\smallskip\noindent\textbf{#1}}}
\newcommand{\alert}[1]{{\color{red}{#1}}}
\newcommand{\sm}{\scriptsize}
\newcommand{\eq}[1]{(\ref{eq:#1})}

\newcommand{\Th}[1]{\textsc{#1}}
\newcommand{\mr}[2]{\multirow{#1}{*}{#2}}
\newcommand{\mc}[2]{\multicolumn{#1}{c}{#2}}
\newcommand{\mca}[3]{\multicolumn{#1}{#2}{#3}}
\newcommand{\tb}[1]{\textbf{#1}}
\newcommand{\ch}{\checkmark}

\newcommand{\red}[1]{{\color{red}{#1}}}
\newcommand{\blue}[1]{{\color{blue}{#1}}}
\newcommand{\green}[1]{\color{green}{#1}}
\newcommand{\gray}[1]{{\color{gray}{#1}}}
\newcommand{\orange}[1]{{\color{orange}{#1}}}
\definecolor{lightgray2}{rgb}{0.9, 0.9, 0.9}

\newcommand{\citeme}[1]{\red{[XX]}}
\newcommand{\refme}[1]{\red{(XX)}}

\newcommand{\fig}[2][1]{\includegraphics[width=#1\linewidth]{fig/#2}}
\newcommand{\figh}[2][1]{\includegraphics[height=#1\linewidth]{fig/#2}}


\newcommand{\tran}{^\top}
\newcommand{\mtran}{^{-\top}}
\newcommand{\zcol}{\mathbf{0}}
\newcommand{\zrow}{\zcol\tran}

\newcommand{\ind}{\mathbbm{1}}
\newcommand{\expect}{\mathbb{E}}
\newcommand{\nat}{\mathbb{N}}
\newcommand{\zahl}{\mathbb{Z}}
\newcommand{\real}{\mathbb{R}}
\newcommand{\proj}{\mathbb{P}}
\newcommand{\prob}{\mathbf{Pr}}
\newcommand{\normal}{\mathcal{N}}

\newcommand{\mif}{\textrm{if}\ }
\newcommand{\other}{\textrm{otherwise}}
\newcommand{\minimize}{\textrm{minimize}\ }
\newcommand{\maximize}{\textrm{maximize}\ }
\newcommand{\st}{\textrm{subject\ to}\ }

\newcommand{\id}{\operatorname{id}}
\newcommand{\const}{\operatorname{const}}
\newcommand{\sgn}{\operatorname{sgn}}
\newcommand{\var}{\operatorname{Var}}
\newcommand{\mean}{\operatorname{mean}}
\newcommand{\trace}{\operatorname{tr}}
\newcommand{\diag}{\operatorname{diag}}
\newcommand{\vect}{\operatorname{vec}}
\newcommand{\cov}{\operatorname{cov}}
\newcommand{\sign}{\operatorname{sign}}
\newcommand{\prj}{\operatorname{proj}}

\newcommand{\sigmoid}{\operatorname{sigmoid}}
\newcommand{\softmax}{\operatorname{softmax}}
\newcommand{\clip}{\operatorname{clip}}

\newcommand{\defn}{\mathrel{:=}}
\newcommand{\peq}{\mathrel{+\!=}}
\newcommand{\meq}{\mathrel{-\!=}}

\newcommand{\floor}[1]{\left\lfloor{#1}\right\rfloor}
\newcommand{\ceil}[1]{\left\lceil{#1}\right\rceil}
\newcommand{\inner}[1]{\left\langle{#1}\right\rangle}
\newcommand{\norm}[1]{\left\|{#1}\right\|}
\newcommand{\abs}[1]{\left|{#1}\right|}
\newcommand{\frob}[1]{\norm{#1}_F}
\newcommand{\card}[1]{\left|{#1}\right|\xspace}
\newcommand{\diff}{\mathrm{d}}
\newcommand{\der}[3][]{\frac{d^{#1}#2}{d#3^{#1}}}
\newcommand{\pder}[3][]{\frac{\partial^{#1}{#2}}{\partial{#3^{#1}}}}
\newcommand{\ipder}[3][]{\partial^{#1}{#2}/\partial{#3^{#1}}}
\newcommand{\dder}[3]{\frac{\partial^2{#1}}{\partial{#2}\partial{#3}}}

\newcommand{\wb}[1]{\overline{#1}}
\newcommand{\wt}[1]{\widetilde{#1}}

\def\xssp{\hspace{1pt}}
\def\ssp{\hspace{3pt}}
\def\msp{\hspace{5pt}}
\def\lsp{\hspace{12pt}}

\newcommand{\cA}{\mathcal{A}}
\newcommand{\cB}{\mathcal{B}}
\newcommand{\cC}{\mathcal{C}}
\newcommand{\cD}{\mathcal{D}}
\newcommand{\cE}{\mathcal{E}}
\newcommand{\cF}{\mathcal{F}}
\newcommand{\cG}{\mathcal{G}}
\newcommand{\cH}{\mathcal{H}}
\newcommand{\cI}{\mathcal{I}}
\newcommand{\cJ}{\mathcal{J}}
\newcommand{\cK}{\mathcal{K}}
\newcommand{\cL}{\mathcal{L}}
\newcommand{\cM}{\mathcal{M}}
\newcommand{\cN}{\mathcal{N}}
\newcommand{\cO}{\mathcal{O}}
\newcommand{\cP}{\mathcal{P}}
\newcommand{\cQ}{\mathcal{Q}}
\newcommand{\cR}{\mathcal{R}}
\newcommand{\cS}{\mathcal{S}}
\newcommand{\cT}{\mathcal{T}}
\newcommand{\cU}{\mathcal{U}}
\newcommand{\cV}{\mathcal{V}}
\newcommand{\cW}{\mathcal{W}}
\newcommand{\cX}{\mathcal{X}}
\newcommand{\cY}{\mathcal{Y}}
\newcommand{\cZ}{\mathcal{Z}}

\newcommand{\vA}{\mathbf{A}}
\newcommand{\vB}{\mathbf{B}}
\newcommand{\vC}{\mathbf{C}}
\newcommand{\vD}{\mathbf{D}}
\newcommand{\vE}{\mathbf{E}}
\newcommand{\vF}{\mathbf{F}}
\newcommand{\vG}{\mathbf{G}}
\newcommand{\vH}{\mathbf{H}}
\newcommand{\vI}{\mathbf{I}}
\newcommand{\vJ}{\mathbf{J}}
\newcommand{\vK}{\mathbf{K}}
\newcommand{\vL}{\mathbf{L}}
\newcommand{\vM}{\mathbf{M}}
\newcommand{\vN}{\mathbf{N}}
\newcommand{\vO}{\mathbf{O}}
\newcommand{\vP}{\mathbf{P}}
\newcommand{\vQ}{\mathbf{Q}}
\newcommand{\vR}{\mathbf{R}}
\newcommand{\vS}{\mathbf{S}}
\newcommand{\vT}{\mathbf{T}}
\newcommand{\vU}{\mathbf{U}}
\newcommand{\vV}{\mathbf{V}}
\newcommand{\vW}{\mathbf{W}}
\newcommand{\vX}{\mathbf{X}}
\newcommand{\vY}{\mathbf{Y}}
\newcommand{\vZ}{\mathbf{Z}}

\newcommand{\va}{\mathbf{a}}
\newcommand{\vb}{\mathbf{b}}
\newcommand{\vc}{\mathbf{c}}
\newcommand{\vd}{\mathbf{d}}
\newcommand{\ve}{\mathbf{e}}
\newcommand{\vf}{\mathbf{f}}
\newcommand{\vg}{\mathbf{g}}
\newcommand{\vh}{\mathbf{h}}
\newcommand{\vi}{\mathbf{i}}
\newcommand{\vj}{\mathbf{j}}
\newcommand{\vk}{\mathbf{k}}
\newcommand{\vl}{\mathbf{l}}
\newcommand{\vm}{\mathbf{m}}
\newcommand{\vn}{\mathbf{n}}
\newcommand{\vo}{\mathbf{o}}
\newcommand{\vp}{\mathbf{p}}
\newcommand{\vq}{\mathbf{q}}
\newcommand{\vr}{\mathbf{r}}
\newcommand{\Vs}{\mathbf{s}}
\newcommand{\vt}{\mathbf{t}}
\newcommand{\vu}{\mathbf{u}}
\newcommand{\vv}{\mathbf{v}}
\newcommand{\vw}{\mathbf{w}}
\newcommand{\vx}{\mathbf{x}}
\newcommand{\vy}{\mathbf{y}}
\newcommand{\vz}{\mathbf{z}}

\newcommand{\vone}{\mathbf{1}}
\newcommand{\vzero}{\mathbf{0}}

\newcommand{\valpha}{{\boldsymbol{\alpha}}}
\newcommand{\vbeta}{{\boldsymbol{\beta}}}
\newcommand{\vgamma}{{\boldsymbol{\gamma}}}
\newcommand{\vdelta}{{\boldsymbol{\delta}}}
\newcommand{\vepsilon}{{\boldsymbol{\epsilon}}}
\newcommand{\vzeta}{{\boldsymbol{\zeta}}}
\newcommand{\veta}{{\boldsymbol{\eta}}}
\newcommand{\vtheta}{{\boldsymbol{\theta}}}
\newcommand{\viota}{{\boldsymbol{\iota}}}
\newcommand{\vkappa}{{\boldsymbol{\kappa}}}
\newcommand{\vlambda}{{\boldsymbol{\lambda}}}
\newcommand{\vmu}{{\boldsymbol{\mu}}}
\newcommand{\vnu}{{\boldsymbol{\nu}}}
\newcommand{\vxi}{{\boldsymbol{\xi}}}
\newcommand{\vomikron}{{\boldsymbol{\omikron}}}
\newcommand{\vpi}{{\boldsymbol{\pi}}}
\newcommand{\vrho}{{\boldsymbol{\rho}}}
\newcommand{\vsigma}{{\boldsymbol{\sigma}}}
\newcommand{\vtau}{{\boldsymbol{\tau}}}
\newcommand{\vupsilon}{{\boldsymbol{\upsilon}}}
\newcommand{\vphi}{{\boldsymbol{\phi}}}
\newcommand{\vchi}{{\boldsymbol{\chi}}}
\newcommand{\vpsi}{{\boldsymbol{\psi}}}
\newcommand{\vomega}{{\boldsymbol{\omega}}}

\newcommand{\rLambda}{\mathrm{\Lambda}}
\newcommand{\rSigma}{\mathrm{\Sigma}}

\newcommand{\vLambda}{\bm{\rLambda}}
\newcommand{\vSigma}{\bm{\rSigma}}

\makeatletter
\newcommand*\bdot{\mathpalette\bdot@{.7}}
\newcommand*\bdot@[2]{\mathbin{\vcenter{\hbox{\scalebox{#2}{$\m@th#1\bullet$}}}}}
\makeatother

\makeatletter
\DeclareRobustCommand\onedot{\futurelet\@let@token\@onedot}
\def\@onedot{\ifx\@let@token.\else.\null\fi\xspace}

\def\eg{\emph{e.g}\onedot} \def\Eg{\emph{E.g}\onedot}
\def\ie{\emph{i.e}\onedot} \def\Ie{\emph{I.e}\onedot}
\def\cf{\emph{cf}\onedot} \def\Cf{\emph{Cf}\onedot}
\def\etc{\emph{etc}\onedot} \def\vs{\emph{vs}\onedot}
\def\wrt{w.r.t\onedot} \def\dof{d.o.f\onedot} \def\aka{a.k.a\onedot}
\def\etal{\emph{et al}\onedot}
\makeatother


\newcommand{\ours}{DToP\xspace}
\newcommand{\Ours}{\emph{deep token pooling} (\ours)\xspace}


\newcommand{\cls}{{\texttt{[CLS]}}\xspace}
\newcommand{\mask}{{\texttt{[MASK]}}\xspace}

\newcommand{\relu}{\operatorname{relu}}
\newcommand{\conv}{\operatorname{conv}}
\newcommand{\aconv}{\operatorname{aconv}}

\newcommand{\fc}{\textsc{fc}}
\newcommand{\gap}{\textsc{gap}}
\newcommand{\bn}{\textsc{bn}}
\newcommand{\dropout}{\textsc{dropout}}

\newcommand{\elm}{\textsc{elm}}
\newcommand{\irb}{\textsc{irb}}
\newcommand{\wav}{\textsc{wb}}
\newcommand{\aspp}{\textsc{aspp}}
\newcommand{\fuse}{\textsc{fuse}}


\def\oxf5k{Ox5k\xspace}
\def\paris6k{Par6k\xspace}
\def\roxf{$\cR$Oxford\xspace}
\def\rox{$\cR$Oxf\xspace}
\def\ro{$\cR$O\xspace}
\def\rpar{$\cR$Paris\xspace}
\def\rpa{$\cR$Par\xspace}
\def\rp{$\cR$P\xspace}
\def\r1m{$\cR$1M\xspace}
\def\rs{$\cR$100k\xspace}

\newcommand\cw{0.8cm}

\newcommand*\circled[1]{\tikz[baseline=(char.base)]{
    \node[shape=circle, draw, inner sep=1pt, 
        minimum height=1pt] (char) {\vphantom{1g}#1};}}


\newcommand{\gain}[1]{{\color{green!60!black}#1}}

\begin{abstract}
   Vision-language models (VLMs) have made significant strides in cross-modal understanding through large-scale paired datasets. However, in fashion domain, datasets often exhibit a disparity between the information conveyed in image and text. This issue stems from datasets containing multiple images of a single fashion item all paired with one text, leading to cases where some textual details are not visible in individual images. This mismatch, particularly when non-co-occurring elements are masked, undermines the training of conventional VLM objectives like Masked Language Modeling and Masked Image Modeling, thereby hindering the model’s ability to accurately align fine-grained visual and textual features. Addressing this problem, we propose Synchronized attentional Masking (SyncMask), which generate masks that pinpoint the image patches and word tokens where the information co-occur in both image and text. This synchronization is accomplished by harnessing cross-attentional features obtained from a momentum model, ensuring a precise alignment between the two modalities. Additionally, we enhance grouped batch sampling with semi-hard negatives, effectively mitigating false negative issues in Image-Text Matching and Image-Text Contrastive learning objectives within fashion datasets. Our experiments demonstrate the effectiveness of the proposed approach, outperforming existing methods in three downstream tasks.
\end{abstract}


    \section{Introduction}
\label{sec:intro}
Recently, there has been rapid progress in developing Vision-Language Pretraining (VLP)~\cite{lu2019vilbert, vlbert, uniter, lxmert, pixelbert, soho, probing, vilt, CLIP, ALIGN, ALBEF, grit-vlp}, paving the way to bridge the gap between visual and textual features. These VLP methods, trained on extensive image-text datasets, have enabled a deeper understanding of semantic alignment across different modalities. By fine-tuning these pretrained models for specific tasks, particularly in data-scarce areas like image-text retrieval, notable performance improvements have been observed. In the pretraining phase, various factors such as model architecture, training objectives, and batch sampling techniques play a crucial role in effectively harnessing the joint representation of multi-modal data.

\begin{figure}[t]
\centering\includegraphics[width=1.0\columnwidth]{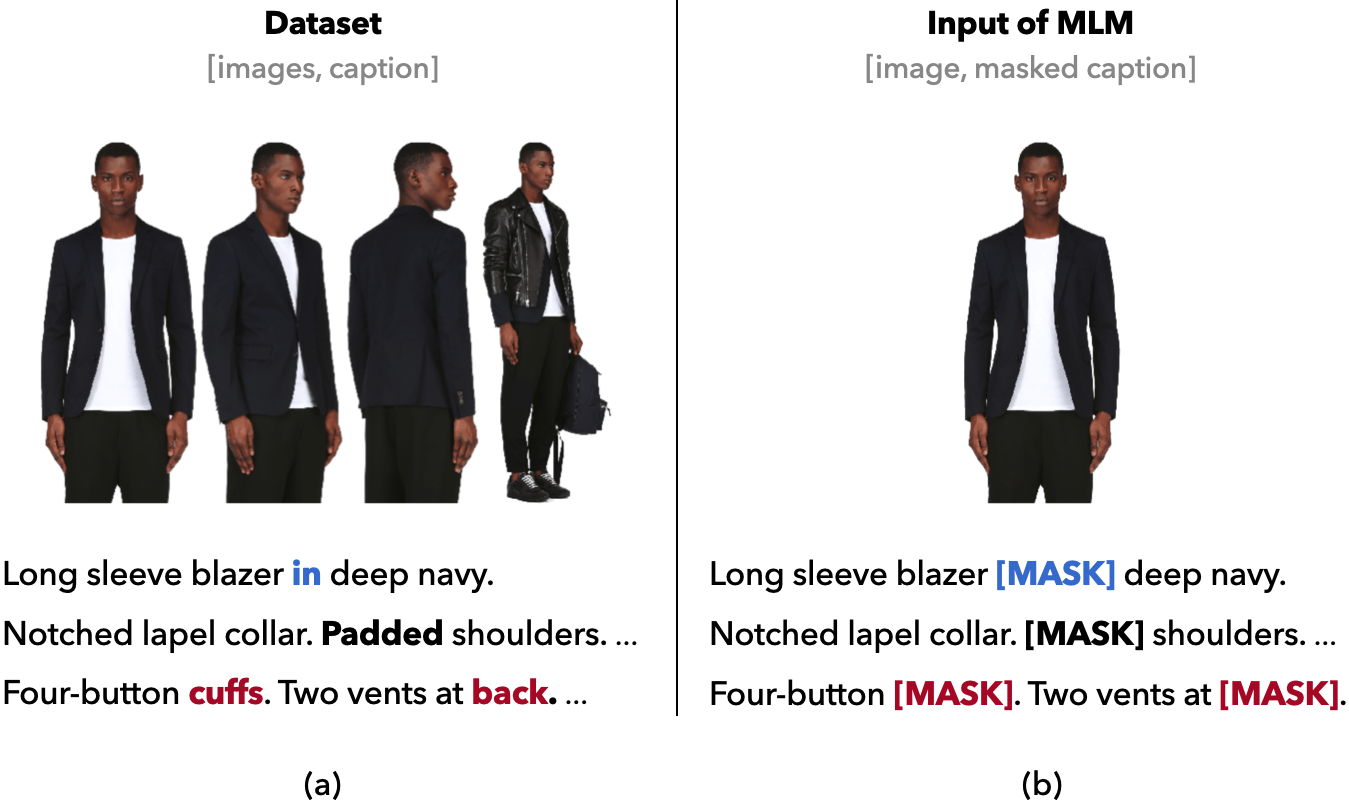}
 \vspace{-8pt}
   \caption{ Example of misaligned masks in the MLM task. }
\label{fig:intro}
 \vspace{-10pt}
\end{figure}

However, there are some issues in applying conventional generic VLP approaches to task-specific domains such as fashion. Fashion VLP models~\cite{fashionbert, zhuge2021kaleidobert, han2022fashionvil, famevil, FashionSAP} typically employ objectives such as Masked Language Modeling (MLM) and Masked Image Modeling (MIM). 
These methods mask elements like text words or image patches, leveraging surrounding context for prediction or reconstruction. They boost cross-modality by making models infer masked text tokens or image patches from aligned features. 
However, existing MLM and MIM often suffer from inherent misalignment limitations because the masks are generated randomly, often leading to unmatched elements being masked.

To illustrate these limitations, consider ~\autoref{fig:intro} (a) shows a single description associated with four images from the FashionGen~\cite{fashiongen} dataset. In ~\autoref{fig:intro} (b), a random masking scenario is shown where the blue \mask might lead the model to predict the masked word using only the text context, thereby not incorporating the image information. Similarly, the two red \mask tokens in (b) lack relevance to the accompanying image, thus hindering the model's ability to connecting between visual and textual features. Furthermore, in the MIM task, random masking may inadvertently cover parts of the image that contain fashion items not described in the text, leading to mismatches during the training of the alignment of cross-modal features. To solve these problem, we propose SyncMask, selecting masks that represent \textit{synchronously co-occurring} features utilizing the vision-language cross-attention map. 


In addition, compared to general domains, the fashion domain often suffers from smaller dataset sizes and less variance in data distribution. This suggests that using standard VLP methods may not adequately distinguish the fine-grained features vital for fashion-related tasks. Thus, we pay attention to a grouped batch sampling technique~\cite{grit-vlp} that similar samples are gradually collected within the batch as the training progresses, impacting the pretraining objectives of Image-Text Contrastive Learning (ITC) and Image-Text Matching (ITM). When similar samples exist within the same batch, it becomes more challenging to differentiate positives and negatives during the training of ITC and ITM compared to using random samples. This encourages the model to more intensely focus on learning fine-grained differences, even with less training.

The existing grouped batch sampling method uses the output features of two uni-modal encoders to find the most similar sample. However, as shown in ~\autoref{fig:intro}, there are many data that have the same caption for multiple images as for the fashion domain. Therefore, a false negative problem that causes actual positive samples in the same batch to be wrongly labelled as negative when learning ITC and ITM arises if the existing methods are used without changes. To overcome this limitation, we propose a semi-hard negative sampling technique with lower similarities between samples that constitute the batch while removing the false negative.

In summary, the main contributions of this study are:
\begin{enumerate}[itemsep=2pt, parsep=0pt, topsep=0pt]
\item \textbf{Synchronized Attentional Masking}: We introduce SyncMask, which replaces random mask in MLM and MIM with targeted mask of co-occurring segments in image-text pairs. By utilizing cross-attention features from a momentum model to generate these masks, this method effectively addresses the problem of mismatched image-text inputs, thereby enhancing fine-grained alignment of cross-modal features.

\item \textbf{Refined Grouped Batch Sampling}: Our method incorporates semi-hard negative sampling to tackle data scarcity and distribution disparities in domain-specific datasets, thereby reducing false negatives.
\end{enumerate}

\section{Related Works}
\label{sec:rworks}

\paragraph{Vision and Language (VL) Model}
    Recently, VLMs have focused on enhancing model architecture and designing objectives to integrate visual and textual features effectively. Early studies~\cite{lu2019vilbert, vlbert, lxmert, uniter} have used object detectors for extracting visual features as an input for a multi-modal transformer along with textual features. The objective for training models extends vanilla BERT~\cite{BERT} to use MLM, MIM, and ITM losses; however, the object detection module incurs a high computational cost for training and inference. Therefore, there have been attempts to replace it with CNN~\cite{pixelbert, soho} or linear projection~\cite{vilt}. These studies commonly train models with MLM and ITM,  tailoring MIM to their specific architectures. Concurrently, CLIP~\cite{CLIP}, ALIGN~\cite{ALIGN} propose models comprising only two unimodal encoders, demonstrating the outstanding representation embedding capabilities of contrastive learning. ALBEF~\cite{ALBEF} add a contrastive learning objective to the previous multi-modal transformer structure for aligning the two modalities before fusion. Based on this model, GRIT-VLP~\cite{grit-vlp} demonstrate improved learning efficiency when configuring batches with hard negative samples. 


\begin{table}
\centering
\small
\setlength{\tabcolsep}{2.9pt}
\begin{tabular}{lcccccc} \toprule
	{\Th{Method}}                        &  {\Th{MM}}    & {\Th{MLM}} & {\Th{MIM}} &  \Th{ATM}   &  \Th{AVM} &  \Th{OnUp} \\ 
  \midrule      
      DMAE~\cite{dmae}                        &            &        &     \ch        &         &      &   \\
      MaskDistill~\cite{peng2022unified}       &           &      &     \ch        &         &   \ch    & \ch   \\
      AttMask~\cite{attmask}                  &                &      &     \ch        &         &   \ch    & \ch   \\
      \midrule      
      ALBEF~\cite{ALBEF}                      &    \ch         &     \ch   &             &         &     &    \\
      MaskVLM~\cite{kwon2023masked}          &   \ch       &   \ch    &      \ch     &         &     &     \\
      MAMO~\cite{mamo}          &   \ch       &   \ch    &      \ch     &         &     &     \\
      \midrule      
      FashionBERT~\cite{fashionbert}          &    \ch         &   \ch      &   \ch       &         &      &   \\
      Kaleido-BERT~\cite{zhuge2021kaleidobert} &    \ch         &   \ch      &   \ch      &   \ch      &   \ch  &     \\
      FashionViL~\cite{han2022fashionvil}     &    \ch         &     \ch   &     \ch        &         &     &    \\
      FashionSAP~\cite{FashionSAP}            &   \ch       &   \ch    &           &         &   &       \\
  \midrule
  \rowcolor{LightCyan}
	 \tb{Ours}                              &       \ch    & \ch       &        \ch     &   \ch    &   \ch & \ch  \\ 
  \bottomrule
\end{tabular}
 \vspace{-8pt}
\caption{Related works \vs Ours on Masked Modeling.  MM:Multi-Modal. MLM:Masked Language Modeling. MIM:Masked Image Modeling. ATM:Attentional Textual Mask. AVM:Attentional Visual Mask. OnUp:Online Update for Attentional Masking}
\label{tab:realated}
 \vspace{-10pt}
\end{table}

\begin{figure*}[t]
\begin{center}
	\fig[1.0]{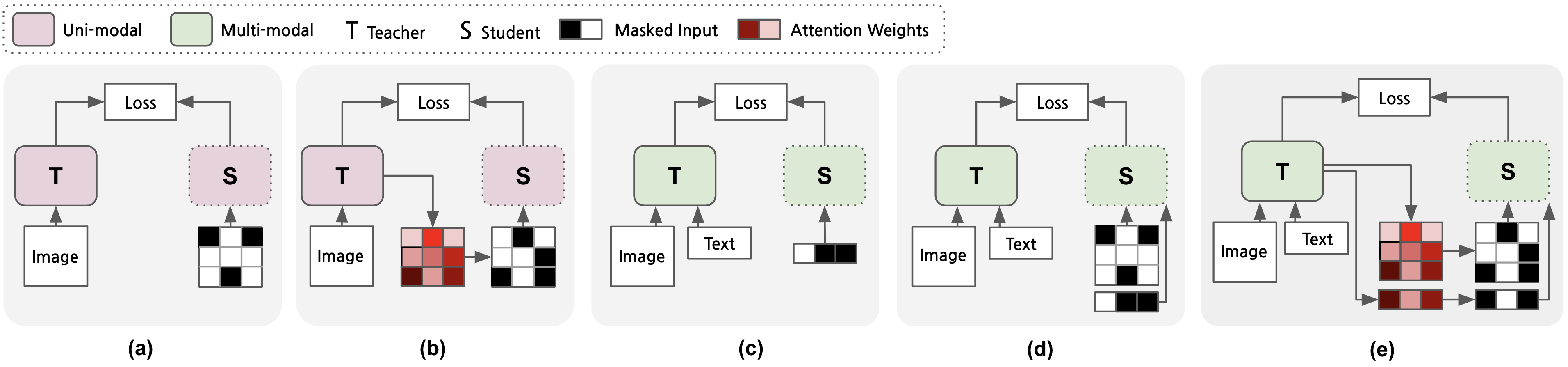}
\end{center}
 \vspace{-8pt}
\caption{Overview of masking strategies using a teacher-student distillation framework. 1) Uni-modal models: (a) random masking, (b) teacher-guided attentional masking. 2) Multi-modal models: (c) random text masking, (d) random image/text masking, (e) teacher-guided cross-attentional masking (Ours).}
\label{fig:Previous}
\vspace{-10pt}
\end{figure*}

\paragraph{FashionVL Model} 
In recent years, various studies~\cite{fashionbert, zhuge2021kaleidobert, han2022fashionvil, famevil, FashionSAP} attempted to capture the finer details of images, building upon established models and pre-training objectives from the general VL task. FashionBERT~\cite{fashionbert} integrates patch-based image features and BERT-based text representations for addressing the limitations of region of interests (RoIs) in capturing fine-grained details. Kaleido-BERT~\cite{zhuge2021kaleidobert} improves fine-grained fashion cross-modality representations through alignment guided masking compared to random masking.  FashionViL~\cite{han2022fashionvil} employs a versatile VLP framework, leveraging two pre-training tasks for capturing the rich fine-grained information of fashion data. FashionSAP~\cite{FashionSAP} employs abstract fashion symbols and an attributes prompt technique for effectively modeling multi-modal fashion attributes. We clarify the importance of the attentional masking technique that employs alignment between MIM and MLM building upon preceding methods. In addition, we underscore a previously unaddressed need for grouped batch sampling within the fashion domain.

\paragraph{Attention-guided Masked Modeling} 
MIM and MLM leverage unmasked contextual clues to predict masked visual and textual elements, respectively. As shown in the upper section of \autoref{tab:realated} and \autoref{fig:Previous} (a, b), in the MIM task, prior studies have evolved from random masking~\cite{mae, beit, simmim, dmae} to attention-guided masking~\cite{li2021mst, peng2022unified, attmask}, demonstrating that targeting highly-attended patches with teacher-student distillation framework improves masked modeling outcomes. Masked modeling also has been pivotal in advancing cross-modal alignment within VLMs, spanning both general~\cite{vlbert,lu2019vilbert, ALBEF, kwon2023masked, mamo} and fashion-specific~\cite{fashionbert, zhuge2021kaleidobert, han2022fashionvil, FashionSAP} domains. This is briefly illustrated in \autoref{fig:Previous} (c, d) and delineated in the middle and lower sections of \autoref{tab:realated}. These enhance the model's proficiency aligning co-occurring visual and textual representations. However, challenges emerge when masking non-co-occurring elements, which hinders the accurate pairing of visual and textual features. Kaleido-BERT~\cite{zhuge2021kaleidobert} improve fine-grained cross-modality representations through text-image alignment-guided masking. This requires additional components for an attention-based alignment generator, increasing computational demands and potential model overfitting on fixed text-image mask pairs. Our approach overcomes these limitations by progressively tailoring masks during end-to-end training.

\begin{figure*}[t]
\begin{center}
	\fig[1.0]{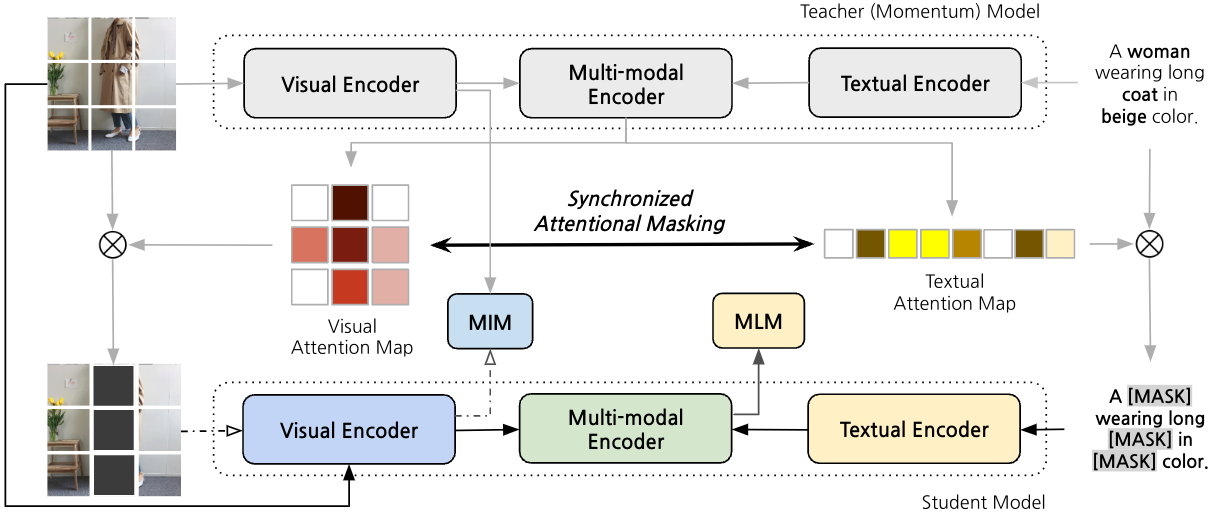}
\end{center}
 \vspace{-4pt}
\caption{ A schematic overview of the SyncMask process: Leveraging cross-attention features from the teacher (momentum) model to generate informative masks for both MIM and MLM tasks. It is important to note that the input for MLM consists of unmasked image paired with masked text. }
\label{fig:ours}
\vspace{-8pt}
\end{figure*}

\section{Methods}
\label{sec:Methods}
We provide an overview of the preliminary aspects, which includes the model architecture and two well-established training objectives for VLP. Subsequently, we present a detailed explanation of the proposed methods, which are synchronized attentional masked modeling and grouped batch sampling with semi-hard negatives.


\subsection{Preliminaries} 

For an image-text pair, we refer input sequences as follows: tokenized text embeddings are denoted as $T = [\vt_\cls, \vt_1, \ldots, \vt_{N}] \in \mathbb{R}^{(N+1) \times D}$ and visual patch embeddings represented as $V = [\vv_\cls, \vv_1, \ldots, \vv_{N'}] \in \mathbb{R}^{(N'+1) \times D}$. Here, $D$, $N$, and $N'$  refer to the transformer dimension, the number of text tokens, and the number of image patches, respectively. Additionally, $\vt_\cls$ and $\vv_\cls$ specifically reference the $\cls$ embeddings. 

The model consists of two components: a teacher model, referred to as the momentum model, and a student model.
The student model, denoted as ${f_{\theta}}$, is parameterized by $\theta$, which includes a textual encoder ${f^T_{\theta}(T)}\in \mathbb{R}^{(N+1) \times D}$, visual encoder ${f^I_{\theta}(V)} \in \mathbb{R}^{(N^{'}+1) \times D}$, and multi-modal encoder ${f^M_{\theta}({f^T_{\theta}(T)},{f^I_{\theta}(V)})}\in \mathbb{R}^{(N+1) \times D}$.
For the momentum model ${f_{{\theta}^{'}}}$, the training parameters are updated by the exponential moving average method, ${\theta}^{'} \leftarrow \beta{\theta}^{'} + (1-\beta){\theta}$, where $\beta$ represents a hyperparameter.
\paragraph{Image-Text Contrastive Learning (ITC)}
     At the front of the multi-modal encoder, ITC pre-aligns the joint latent space of the textual encoder and visual encoder. This objective have proved its effectiveness in VLMs~\cite{CLIP, ALBEF, BLIP, grit-vlp, han2022fashionvil, FashionSAP}. We also adopt the ITC loss framework proposed by~\cite{ALBEF, BLIP, FashionSAP}, which incorporates a momentum encoder for utilizing soft labels as ITC training targets, thereby addressing potential positive instances within negative pairs.
\paragraph{Image-Text Matching (ITM)}
     For the ITM loss, the model classifies image-text pairs as either matched (positive) or not matched (negative) using a joint representation obtained from the $\cls$  token output embedding of the multi-modal encoder. This vector is passed through an FC layer and softmax for binary prediction. Like ALBEF~\cite{ALBEF}, we exploit hard negatives in the ITM task, identifying pairs that share similar semantics but differ in fine-grained details, using in-batch contrastive similarity from ITC.

\label{sec:Preliminaries}

\subsection{Synchronized Attentional Masked Modeling}
\label{sec:syncedTextVisual}
We extend the use of momentum model, a self-supervised tool for momentum distillation outlined in MoCo~\cite{MoCo} and ALBEF~\cite{ALBEF}, beyond its conventional role of generating pseudo-labels. We employ its multi-modal encoder, which calculate the cross-attention map to identify patches and tokens where image and text features strongly correlate. These elements, indicated by heightened attention weights, are then selectively masked. This attentional masking, targeting \textit{synchronously co-occurring} features, enhances cross-modality over traditional random masking methods in the MLM and MIM phases. Moreover, the momentum model's output features provide enhanced labels for these masked regions, offering a depth of information beyond conventional discrete labels. We will explore this method further to understand its full potential in capturing intricate multi-modal interactions.
\paragraph{Vision-Language Synchronized Attentionl Masking}
MLM and MIM elevate the alignment of visual and textual representations in models. However, when non-co-occurring elements are masked, these techniques face limitations that restrict the model's ability to accurately match multi-modal features.
To address this issue, we propose a Synchronized attentional Masking (SyncMask) strategy into masked multi-modal modeling objectives. As depicted in \autoref{fig:ours}, we extracted two sets of synchronized attention weights from the cross-attention module of the multi-modal encoder's last layer in the momentum model. The module enables the model to fuse image-text features using a cross-attention mechanism that processes a query ($Q^T$), key ($K^I$), and value ($V^I$), as represented by the following equation:

 \vspace{-5pt}
\begin{equation}
\label{eq:textMHA}
   \mathtt{Attention}{(Q^{T}_i, K^{I}_i, V^{I}_i)} = {\alpha}(Q^{T}_i, K^{I}_i) \odot V^{I}_i
\end{equation}

where $1 \leq i \leq H$, with $H$ denoting the number of heads in the MHA, and ${\odot}$ representing the Hadamard product. In this context, $\alpha$ refers the cross-attention function, which can be expressed as:
 \vspace{-5pt}
\begin{equation}
\label{eq:attweight_text}
    \alpha(Q^{T}_i, K^{I}_i) = \mathtt{softmax}{(\frac{Q^{T}_{i}(K^{I}_i)^\top}{\sqrt{d}})} \in \mathbb{R}^{(N'+1) \times (N+1) }
\end{equation}
The function $\alpha(Q^{T}_i, K^{I}_i)$ computes the attention weights for the image from the perspective of the text. Similarly, by altering the query ($Q$) and key ($K$), we calculate the text attention weight from the image perspective, as represented by the following equation:
 \vspace{-5pt}
\begin{equation}
\label{eq:attweight_image}
    \alpha(Q^{I}_i, K^{T}_i) = \mathtt{{softmax}}{(\frac{Q^{I}_{i}(K^{T}_i)^\top}{\sqrt{d}})}  \in \mathbb{R}^{(N+1) \times (N'+1)}
    \end{equation}


Utilizing ~\autoref{eq:attweight_text} and ~\autoref{eq:attweight_image}, we derive two synchronized textual-visual cross-attention weights. 
These weights, $\vo^T \in \mathbb{R}^{N}, \vo^I \in \mathbb{R}^{N'},$ are obtained by averaging the patch tokens of the last layer, excluding the \cls token. 
This process allows us to map each word in the sentence sequence to its corresponding attention in $\vo^T$. Further, $\vo^I$ can be reshaped to $\mathbb{R}^{P \times P}$, which aligns with the image patches.
 \vspace{-5pt}
\begin{align}
  \mathtt{Idx}^{T} = \mathtt{shuffle}(\mathtt{sort}^{\mathrel{desc}}(\vo^T)[\geq L])[\geq K]
\label{eq:sorting_weight_text}
\end{align}
 \vspace{-10pt}
\begin{align}
  \mathtt{Idx}^{I} = \mathtt{shuffle}(\mathtt{sort}^{\mathrel{desc}}(\vo^I)[\geq L'])[\geq K']
\label{eq:sorting_weight_image}
\end{align}

In ~\autoref{eq:sorting_weight_text} and ~\autoref{eq:sorting_weight_image}, we first sort the attention weights $\vo^T$ and $\vo^I$ in descending order ($\mathtt{sort}^{desc}$), and then extract their indices ($\mathtt{Idx}$). ~\autoref{eq:sorting_weight_text} focuses on indices corresponding to the top $L$ values of $\vo^T$, where $L$ is a threshold greater than the actual mask size $K$. These indices are randomly shuffled ($\mathtt{shuffle}$) to introduce randomness, and ultimately, only those indices that satisfy the condition $K \leq L$ are retained. A similar approach is applied to $K'$ and $L'$ in ~\autoref{eq:sorting_weight_image}. The parameters $K, K', L,$ and $L'$ are defined based on a mask ratio $r \in [0, 1]$.

The final attention masks for textual and visual components are represented by the vectors $\vm^{T}$ and $\vm^{I}$, respectively. The computation of these masks utilizes the indices $\mathtt{Idx}^{T}$ for text and $\mathtt{Idx}^{I}$ for images, as formulated in the equation below:
\begin{equation}  
\label{eq:masked}
\vm^{T}[\mathtt{Idx}^{T}] \leftarrow 1, \quad  \vm^{I}[\mathtt{Idx}^{I}] \leftarrow 1
\end{equation}

\begin{figure}[t]
\centering\includegraphics[width=1.0\columnwidth]{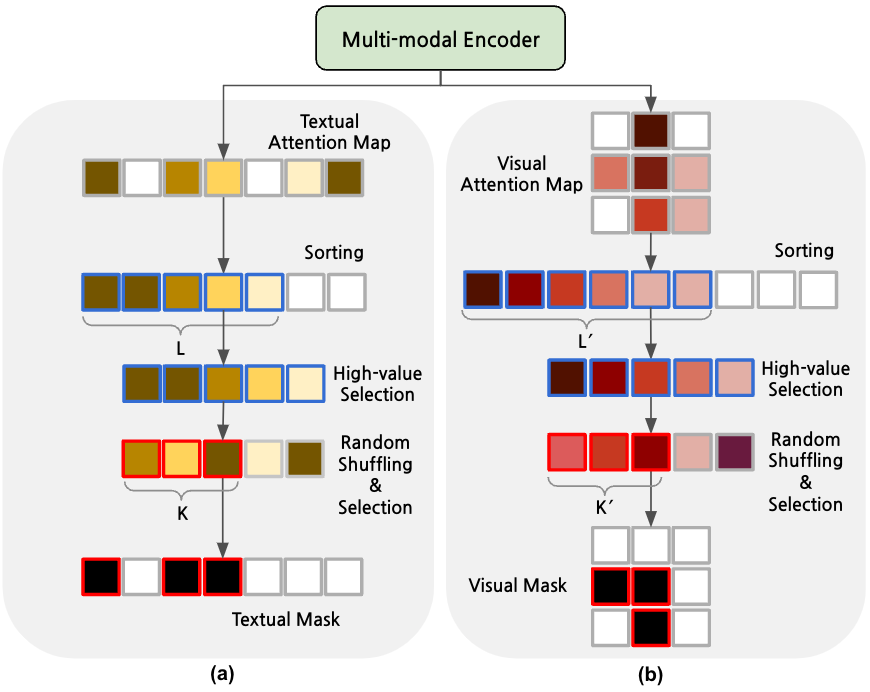}
 \vspace{-4pt}
   \caption{ Selection phase of the SyncMask }
\label{fig:mask}
\end{figure}

Initially, the vectors $\vm^{T}$ and $\vm^{I}$ are zero-initialized with dimensions $N$ (for text) and $N'$ (for images). They are then updated to binary attention masks, which are integral in capturing the synchronized interaction between textual and visual elements, as illustrated in \autoref{fig:mask}. Subsequently, these masks are employed in the masked modeling processes for text and images (detailed in \autoref{eq:mask_text_1} and \autoref{eq:mask_image_1}).

Building upon this foundation, the text mask vector $\vm^{T}$ consists of elements $[m^t_{1}, \dots, m^t_{N}] \in \{0, 1\}^N$, while the image mask vector $\vm^{I}$ is composed of $[m^i_{1}, \dots, m^i_{N'}] \in \{0, 1\}^{N'}$. For each tokenized text vector, the masked version $\tilde{\mathbf{t}}_i$ is determined by:
\begin{equation}\label{eq:mask_text_1}
    \tilde{\mathbf{t}}_j = (1 - m^t_{j}) \cdot \mathbf{t}_j + m^t_{j} \cdot \mathbf{t}_{\text{mask}}
\end{equation}
In this equation, \(1 \leq j \leq N\), and $\mathbf{t}_{\text{mask}}$ denotes the special token used for textual masking. In the context of Masked Image Modeling (MIM), each image patch is processed with a learnable mask, resulting in the masked image vector $\tilde{\mathbf{v}}_k$:
\begin{equation}\label{eq:mask_image_1}
    \tilde{\mathbf{v}}_k = (1 - m^i_{k}) \cdot \mathbf{v}_k + m^i_{k} \cdot \mathbf{v}_{\text{mask}}
\end{equation}

Here, \(1 \leq k \leq N'\), and $\mathbf{v}_{\text{mask}}$ represents the learnable mask embedding~\cite{dino}. The masked tokenized inputs are thus represented as $\tilde{T} = [\mathbf{t}_{\text{cls}}; \tilde{\mathbf{t}}_1; \dots; \tilde{\mathbf{t}}_{N}]$ for text and $\tilde{V} = [\mathbf{v}_{\text{cls}}; \tilde{\mathbf{v}}_1; \dots; \tilde{\mathbf{v}}_{N'}]$ for image. These masked inputs are processed using the proposed SyncMask $\vm^T$ and $\vm^I$.

\begin{table*}
  \centering
  \setlength\tabcolsep{12pt}
  \begin{tabular}{cccccccc}
    \toprule
    \multirow{2}{*}{\Th{Methods}} & \multicolumn{1}{c}{I2T$^{\star}$} & \multicolumn{1}{c}{T2I$^{\star}$} & \multirow{2}{*}{\Th{Mean}$^{\star}$} & \multicolumn{1}{c}{I2T} & \multicolumn{1}{c}{T2I} & \multirow{2}{*}{\Th{Mean}}\\
    \cmidrule(r){2-3} \cmidrule(r){5-6}
    ~ & R@1 & R@1 &  & R@1 & R@1  & \\
    \midrule
    VSE++\cite{vse++}     & 4.59 &  4.60 &  4.60 & -- & -- & -- \\
    VL-BERT~\cite{vlbert} & 19.26 & 22.63 & 20.95 & -- & -- & --\\
    ViLBERT~\cite{lu2019vilbert} & 20.97 & 21.12 & 21.05 & -- & -- & --\\
    Image-BERT~\cite{qi2020imagebert} & 22.76 & 24.78 & 23.77 & -- & -- & --\\
    OSCAR~\cite{li2020oscar} & 23.39 &  25.10 & 24.25 & -- & -- & -- \\
    FashionBERT~\cite{fashionbert} & 23.96  & 26.75  & 25.36 & -- & -- & --\\
    Kaleido-BERT~\cite{zhuge2021kaleidobert} & 27.99 & 33.88 & 30.94 & -- & -- & --\\    
    CommerceMM~\cite{yu2022commercemm} & 41.60  & 39.60 & 62.75 & -- & -- & -- \\
    EI-CLIP~\cite{EICLIP} & 38.70 & 40.06 & 39.38 &  25.70 & 28.40 & 27.05\\ 
    ALBEF~\cite{ALBEF} & 63.97 & 60.52 & 62.20 & 41.68 & 50.95 & 46.32 \\
    FashionViL~\cite{han2022fashionvil} & 65.54 & 61.88 & 63.71 & 42.88 & 51.34 & 47.11 \\    
    FashionSAP~\cite{FashionSAP} & {73.14} & {70.12} & {71.63}  & {54.43} & {62.82}  & {58.63}\\
    \midrule
    \rowcolor{LightCyan}
    Ours & \textbf{75.00}  & \textbf{71.00} & \textbf{73.00}  & \textbf{55.39} & \textbf{64.06} &  \textbf{59.73}\\
    \bottomrule
  \end{tabular}
  \vspace{-5pt}
  \caption{Cross-modal retrieval result on FashionGen\cite{fashiongen} in the sub/full set of evaluation following previous work. $\star$:sub set.}
  \label{tab:retrievalsub}
   \vspace{-5pt}
\end{table*}

  \paragraph{Synchronized Attentional Masked Language Modeling}

MLM predicts masked words based on the surrounding contextual text and image. Many existing VLMs applied the MLM method proposed in BERT~\cite{BERT}, randomly masking words with a probability of 15\%. However, this approach may not be suitable for vision-language datasets with short caption lengths, especially for nonstandard datasets such as fashion, demanding a thorough understanding of fine-grained attributes. We leverage previous works that addressed these issues by increasing masking probabilities~\cite{grit-vlp} and employing attribute prompts~\cite{FashionSAP}. Building upon these methods, we employ masks generated by SyncMask which is contextually attuned to the corresponding image. 

Let $\bm{h}(V, \tilde{T})$ denote the model's predicted probability for a masked token. $\tilde{\bm{y}}$ denote a one-hot vocabulary distribution in which the ground-truth token is assigned a probability of 1. MLM minimize cross-entropy loss, described as follows:



\vspace{-5pt}
\begin{equation}
\label{eq:MLM_loss}
   \mathcal{L}_\Th{MLM} = \mathbb{E}_{(V,\tilde{T}) \sim D} [\Th{CE}( \tilde{\bm{y}}, \bm{h}(V, \tilde{T})], 
\end{equation}
where $\Th{CE}(\cdot, \cdot)$ refers the cross-entropy between two vectors.  
  \paragraph{Synchronized Attentional Masked Image Modeling}
  In the distillation-based MIM~\cite{attmask,li2021mst, peng2022unified}, a teacher encoder sees the full image, whereas the student encoder, seeing the masked image, tackles the reconstruction objective. Our method adopts a similar objective framework, but with a key difference: our masks, generated through SyncMask, are designed to reflect text-informed elements in masked image. To calculate the MIM loss, the following approach is used:
 \vspace{-10pt}
\begin{multline}
\label{eq:DIST}
\Th{Dist}( \bm{f}_{\theta'}^{I}(V),\, \bm{f}^{I}_{\theta}(\tilde{V})) \\
= \frac{1}{\Omega(\vm^{I})} \sum_{k=1}^{N'} \vm^{I}_k \cdot \ell_1^{\mathrel{Smooth}}(\bm{f}_{\theta'}^{I}(V)_k, \bm{f}^{I}_{\theta}(\tilde{V})_k)
\end{multline}
where $\bm{f}_{\theta'}^{I}(\cdot)_k$ and $\bm{f}^{I}_{\theta}(\cdot)_k$ represent the output feature of teacher and student model for the $k$-th image patch, respectively. $\Omega(\cdot)$ means the number of elements with a value of 1 in a vector.
 \vspace{-5pt}
\begin{equation}
\label{smoothL1}
  \ell_1^{\mathrel{Smooth}} (a,b) =
  \begin{cases}
    0.5(a-b)^2& \text{if } |a-b| < \gamma\\
    |a-b| - 0.5 & \text{otherwise},
  \end{cases}
\end{equation}

where $\ell_1^{\mathrel{Smooth}}$~\cite{fastrcnn} represents a robust L1 loss less sensitive to outliers than the L2 loss and $\gamma$ is a hyperparameter set to 1. Conclusively, the training objective of MIM can be formulated as:

 \vspace{-10pt}
\begin{equation}
\label{eq:MIM_loss}
\mathcal{L}_\Th{MIM} = \mathbb{E}_{(V,\tilde{V}) \sim D} [\Th{Dist}( \bm{f}_{\theta'}^{I}(V),\, \bm{f}^{I}_{\theta}(\tilde{V}))]
\end{equation}
The final loss ($\mathcal{L}$) is given as:
\begin{equation}
\label{eq:loss}
   \mathcal{L} = \mathcal{L}_\text{MIM} + \mathcal{L}_\text{MLM} + \mathcal{L}_\text{ITC} + \mathcal{L}_\text{ITM}      
\end{equation}
where $\mathcal{L}_\text{ITC}$ and $\mathcal{L}_\text{ITM}$ denote ITC and ITM, respectively. Due to space constraints, detailed formulations of these two losses are elaborated in the Appendix.


\subsection{Grouped Batch with Semi-hard Negatives}
\label{sec:grit}
In the generic domain,~\cite{grit-vlp} proposed the GRIT strategy for enhancing training effectiveness by forming mini-batches with similar examples. In this strategy, the \textit{grouping based on similarity} phase plays a crucial role. During this phase, similarity calculations are performed in both directions, utilizing the $\cls$ outputs from the unimodal encoders. For each example, the algorithm iteratively identifies the index with the highest similarity, alternating between the image-to-text and text-to-image directions in a sequential manner. These highly similar indices are grouped together within the mini-batch, ensuring that each mini-batch consists of examples exhibiting the highest possible similarity.

However, in fashion datasets, this strategy leads to a false-negative problem in ITM and ITC, where true positives are mislabeled as negatives within mini-batches. To address this problem, we opt to group semi-hard negatives with relatively lower similarity (the $s^{th}$ highest pairwise similarity) instead of the highest ($1^{st}$) during the \textit{grouping based on similarity} phase, where $s$ represents a predefined hyperparameter that is greater than 1. In addition, we prevent true positive samples from grouping by considering the item indices of the samples. Through this approach, the model is trained with negatives that are similar but exhibit meaningful differences, thereby enabling the effective learning of fine-grained distinctions with a limited dataset. More details are in the Appendix.

\begin{figure*}[t]
\begin{center}
	\fig[1.0]{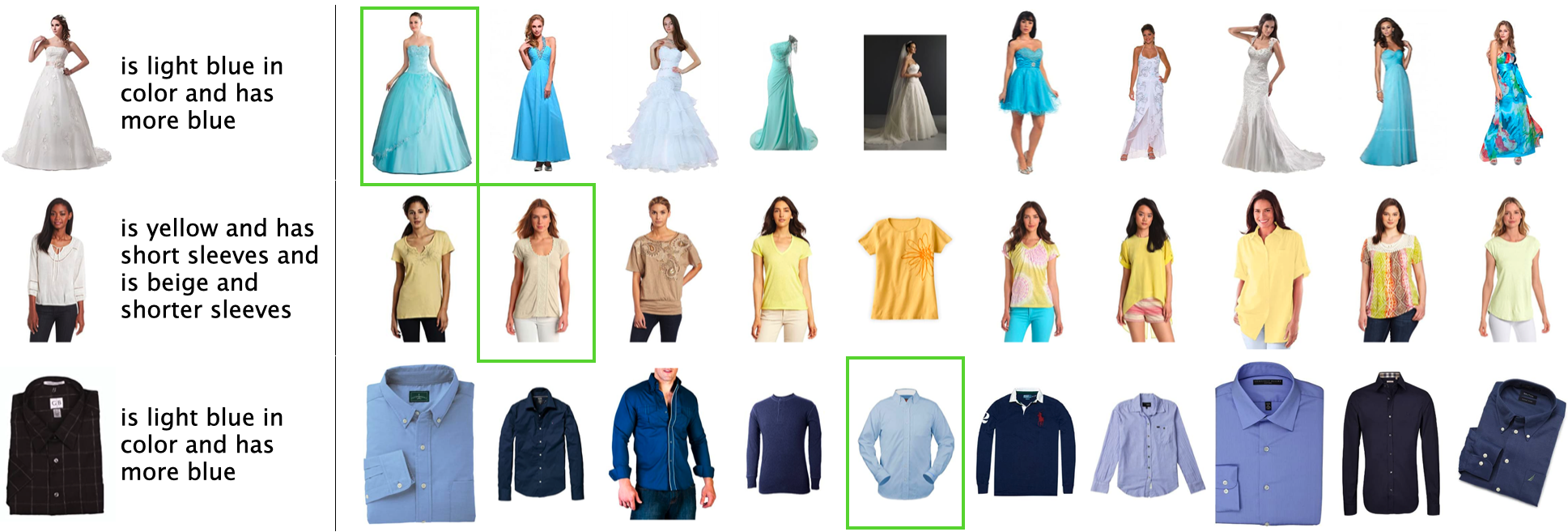}
\end{center}
\vspace{-10pt}
\caption{The top-10 TGIR results of the SyncMask model on the FashionIQ dataset. On the left, the reference images paired with their guided descriptions are shown, while the right side presents the model's predicted images ranked by descending scores. Ground truth images are distinctly outlined with a green bounding box. It is worth mentioning that the set of predictions includes other images that also qualify as suitable matches.}
\label{fig:tgir_example}
\end{figure*}

\begin{table*}
  \centering
    \setlength\tabcolsep{10pt}
  \begin{tabular}{ccccccccc}
    \toprule
    \multirow{2}{*}{\Th{Methods}} & \multicolumn{2}{c}{\Th{Dress}} & \multicolumn{2}{c}{\Th{Toptee}} & \multicolumn{2}{c}{\Th{Shirt}} & \multicolumn{2}{c}{\Th{Mean}}\\
    \cmidrule(r){2-3} \cmidrule(r){4-5} \cmidrule(r){6-7} \cmidrule(r){8-9}
    ~ & R@10 & R@50 & R@10 & R@50 & R@10 & R@50 & R@10 & R@50 \\
    \midrule
    CIRR\cite{cirr} &17.45 &40.41 &21.64 &45.38 &17.53 &38.81 &18.87 &41.53 \\
    VAL\cite{val}  & 22.53 & 44.00 & 27.53 & 51.68  & 22.38 & 44.15 & 24.15 &46.61 \\
    CosMo\cite{lee2021cosmo} & 25.64 & 50.30 & 29.21 & 57.46  & 24.90 & 49.18 &26.58 &52.31 \\
    DCNet\cite{2021dcnet} & 28.95 &56.7 &30.44 &58.29 &23.95 &47.3 &27.78 &54.10 \\
    FashionVLP\cite{goenka2022fashionvlp} & 32.42 & 60.29 & 38.51 & 68.79 & 31.89 & 58.44 & 34.27 & 62.51 \\
    FashionViL\cite{han2022fashionvil} & 33.47 & 59.94  & 34.98 & 60.79&  25.17 & 50.39 & 31.21 & 57.04\\    
    FashionSAP~\cite{FashionSAP} & 33.71 & 60.43 & 41.91 & 70.93 & 33.17 & 61.33 & 36.26 & 64.23\\
    \midrule
     \rowcolor{LightCyan}
     Ours & \textbf{33.76} & \textbf{61.23}  & \textbf{44.82} & \textbf{72.06} &  \textbf{35.82} & \textbf{62.12} & \textbf{38.13} & \textbf{65.14} \\
    \bottomrule
  \end{tabular}
   \vspace{-7pt}
  \caption{Text-guided image retrieval performance in FashionIQ\cite{fashioniq}}
  \label{tab:tmir}
  \vspace{-5pt}
\end{table*}

\section{Experiments}
\label{sec:Experiments}
\subsection{Implementation Details}
\label{sec:Implementation}
The foundational architecture of the proposed model is aligned with prior works for demonstrating the effectiveness of the proposed approach~\cite{ALBEF, grit-vlp, FashionSAP}. The image encoder adopts the architecture of ViT-B16~\cite{vit}, whereas the text encoder comprises the first six blocks of the BERT-bas3~\cite{BERT}. The multi-modal encoder extends the self-attention layers of the last six blocks of BERT with the cross-attention layers. The proposed model is initialized with pre-trained ALBEF~\cite{ALBEF} same as the Fashion-SAP to ensure a fair comparison~\cite{FashionSAP}. 
In addition, we employ the same data augmentation and prompt input strategies as FashionSAP~\cite{FashionSAP}. During pre-training, we conduct experiments using 8 RTX 3090 GPUs each with a batch size of 8 for 30 epochs. We adopt a momentum queue size of 48,000 to facilitate grouped batch sampling, which is consistent with GRIT-VLP~\cite{grit-vlp}. The input image size is set to 256 $\times$ 256. We apply the AdamW~\cite{adamw} optimizer with a learning rate of 6e-5.

\subsection{Datasets}
\label{sec:Datasets}
\paragraph{FashionGen~\cite{fashiongen}} FashionGen comprises 320K text-image pairs and 40K unique fashion items, each represented by multiple images from different angles. For pre-training, we employ the FashionGen train set, which contains approximately 260.5K text-image pairs. In addition, FashionGen supports various downstream tasks, including text-to-image retrieval, image-to-text retrieval, category recognition, and subcategory recognition.
 \vspace{-8pt}
\paragraph{FashionIQ~\cite{fashioniq}} FashionIQ encompasses 77K unique fashion items and includes 18K training triplets (i.e., query image, modified text, target image) and 6K validation datasets for a text-guided image retrieval task. It contains three different categories: Dress, Toptee, and Shirt.

\subsection{Downstream Tasks}
\label{sec:Downstreamtasks}
      \paragraph{Cross-modal Retrieval} We evaluate a cross-modal retrieval task that includes image-text retrieval (ITR) and text-image retrieval (TIR). Further, ITR focuses on finding relevant textual descriptions for a given image query. TIR, the inverse task, retrieves pertinent images based on a textual query. These tasks assess the effectiveness of the model in capturing cross-modal relationships between text and images within retrieval scenarios. Following the previous works~\cite{han2022fashionvil, FashionSAP}, we evaluate cross-modal retrieval not only on the subset with 1K retrievals but also on the full dataset of FashionGen~\cite{fashiongen}. The results, including R@1 scores for both subset and the full set, are presented in Table \autoref{tab:retrievalsub}, demonstrating an improved performance compared to that of the previous results.
    \paragraph{Text-guided Image Retrieval} This task aims to retrieve target images by considering query pairs that reference image and modified descriptions; this is more challenging than traditional retrievals. Therefore, we need to select the target image for identifying minor differences in the changes in description while maintaining the characteristics of the reference image. For a fair comparison, we adopt a similar fine-tuning as outlined in the FashionSAP~\cite{FashionSAP}. In the actual dataset, there are many images that match with the query pairs in addition to the target image referred to as the real correct answer, as shown in \autoref{fig:tgir_example}. Thus, a qualitative evaluation can be conducted in that the model finds similar images well in addition to the actual correct answer. As shown in \autoref{tab:tmir}, the proposed model surpasses previous models and demonstrates state-of-the-art performance.
%
\begin{table}
  \centering \small
  \setlength\tabcolsep{7.2pt}
  \begin{tabular}{ccccc}
    \toprule
    \multirow{2}{*}{\Th{Methods}} & \multicolumn{2}{c}{CR} & \multicolumn{2}{c}{SCR}\\
    \cmidrule(r){2-3} \cmidrule(r){4-5}
    ~ & Acc & Macro-F & Acc & Macro-F \\
    \midrule
    F-BERT\cite{fashionbert} &91.25 & 70.50 & 85.27& 62.00\\
    K-BERT\cite{zhuge2021kaleidobert} & 95.07 & 71.40 & 88.07 & 63.60\\
    F-ViL\cite{han2022fashionvil} & 97.48 & 88.60 & 92.23 & 83.02  \\
    FashionSAP~\cite{FashionSAP} & 98.34 & {89.84} & \textbf{94.33} & 87.67 \\
     \midrule
    \rowcolor{LightCyan}
    Ours & \textbf{98.41} & \textbf{90.31} & 94.21 & \textbf{87.83}  \\  
    \bottomrule
  \end{tabular}
  \vspace{-7pt}
  \caption{CR and SCR results on FashionGen\cite{fashiongen}.}
  \label{tab:catereg}
\end{table}
%
    \paragraph{Category / Subcategory Recognition} In this downstream task, the objective is to classify the category and subcategory of fashion items using the textual and visual information provided. In line with earlier studies~\cite{fashionbert, zhuge2021kaleidobert, han2022fashionvil, FashionSAP}, we simply attach a linear layer to the $\cls$ token, which serves as the fusion feature, for task label prediction. As indicated in \autoref{tab:catereg}, our proposed model exhibits competitive performance compared to existing models.


\subsection{Ablation Study}
\label{sec:Ablation}

\begin{figure}[t]
\centering\includegraphics[width=1.0\columnwidth]{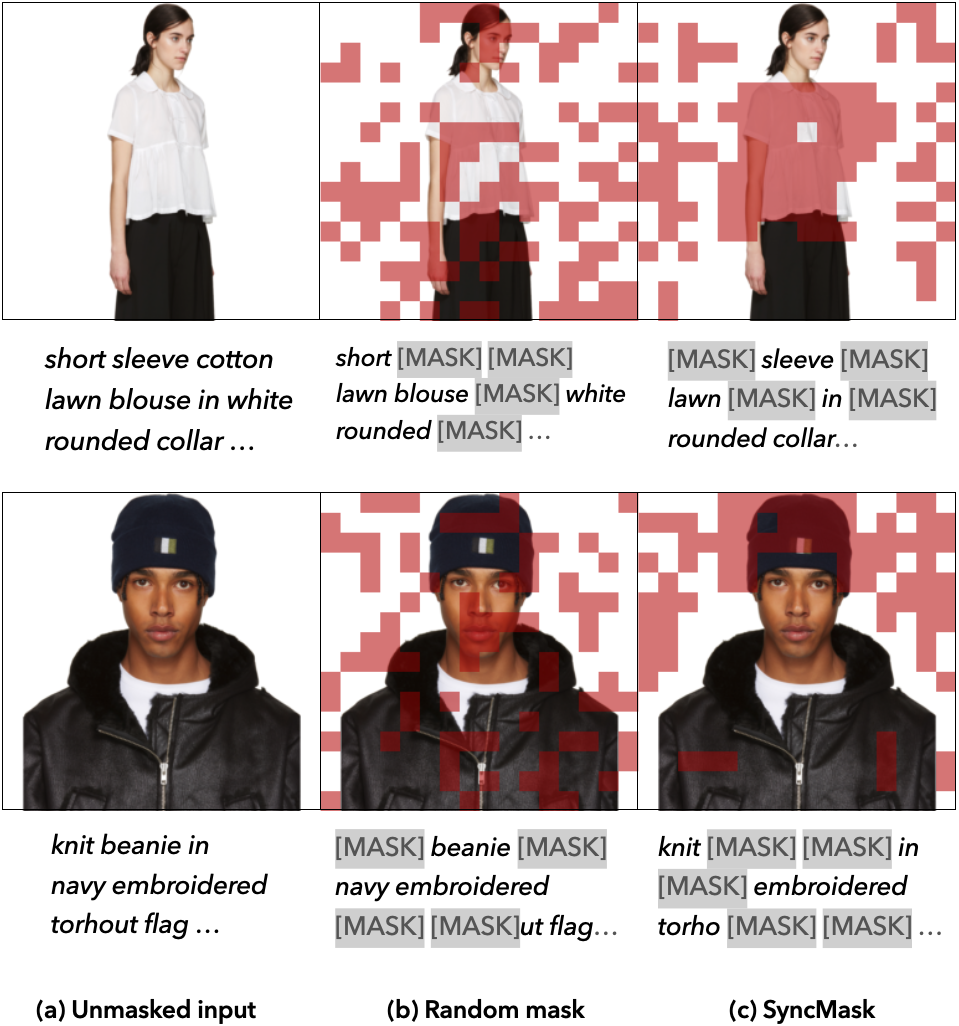}
    \vspace{-4pt}
   \caption{ Examples of random masking (b) and SyncMask (c) for MIM and MLM. The latter applies masks to pertinent features in both the input image and text (a), offering a more context-sensitive selection than the former. }
\label{fig:mask_example}
\end{figure}

\paragraph{Random \vs Attentional Masking}
We ablate the SyncMask with other non-synchronized masking methods. \autoref{tab:Ablation} lists the comparison results for the combination of random and attentional masking. In this experiment, we evaluated five experiments for I2T (R@1), T2I (R@1), CR (Macro-F), SCR (Macro-F), and TMIR (R@10), which were evaluated in the previous experiments. In these experiments, the results of the three experiments with at least one attentional masking were higher than those with no attentional masking (all random). This can be considered an indication of the efficacy of attentional masking. Further, the synergistic performance of the proposed synchronized textual-visual attentional masking (\gain{+2.14}) is greater than that of only one attentional masking (\gain{+0.4}, \gain{+0.66}), which shows the superiority of the SyncMask approach. \autoref{fig:mask_example} displays the distinction between applying random masks (b) and SyncMask (c) to unmasked image-text pairs (a) from the FashionGen~\cite{fashiongen}. SyncMask strategically places masks on image patches that correlate with the text and applies masks to the text informed by the image details, unlike random masking.
\paragraph{Random \vs Hardest \vs Semi-hard Negative Sampling} 
\autoref{tab:Ablation2} compare the effectiveness of various grouped mini-batch sampling strategies across five downstream tasks. The methods evaluated encompassed four scenarios: random grouping, hardest grouping (\gain{+1.90}), hardest grouping with exclusion of false negatives (\gain{+1.98}), and semi-hard grouping also excluding false negatives (\gain{+2.46}). These findings suggest that grouping similar samples in a mini-batch is more beneficial for learning than composing batches with random samples. However, given that the fashion dataset often has multiple captions per image or vice versa, performance gains were observed when systematically preventing the grouping of false negatives by using indexing. Nevertheless, due to the prevalence of inherently similar samples that cannot be systematically excluded, we opted for grouping semi-hard negatives instead of the hardest ones, which resulted in a significant performance boost. This highlights the importance of further research from a data perspective, not just in terms of model architecture or loss functions.

\begin{table}
  \centering \scriptsize
  \setlength\tabcolsep{2.8pt}
  \begin{tabular}{ccccccccccccccc}
    \toprule
    \multirow{2}{*}{\Th{$R$t}} & \multirow{2}{*}{\Th{$R$v}}& \multirow{2}{*}{\Th{$A$t}}  & \multirow{2}{*}{\Th{$A$v}} & \multirow{2}{*}{\Th{Mean}} & \multirow{2}{*}{\Th{Gain}} & \Th{I2T} & \Th{T2I} & \Th{CR} & \Th{SCR} & \Th{TMIR}\\
    ~ &~ & ~ & ~& ~& ~& R@1 &R@1 &Macro-F &Macro-F &R@10\\
    \midrule
   \checkmark & \checkmark & ~ & ~ & 70.31 & ~ & 73.30 & 69.80 & 86.21  & 86.16  & 36.08  \\ 
   & \checkmark  & \checkmark & ~ & 70.71 & \gain{+0.40} & 74.80 & 70.10 &  86.49 & 86.31 & 35.84\\ 
   \checkmark  &  & & \checkmark & 70.97  & \gain{+0.66} &74.30 & 70.80 & 85.12 & 86.78 & 37.87 \\
   \rowcolor{LightCyan}
     &   & \checkmark & \checkmark & \textbf{72.45} & \gain{+2.14} & \textbf{75.00} & \textbf{71.00} & \textbf{90.31} & \textbf{87.83} & \textbf{38.13} \\
    \bottomrule
  \end{tabular}
  \vspace{-8pt}
  \caption{Ablation study results for \emph{Random} \vs \emph{Attention} Masked Modeling on five downstream tasks. $R$t:Random Textual masking. $R$v:Random Visual masking. $A$t:Attentional Textual masking. $A$v:Attentional Visual masking.}
  \label{tab:Ablation}
\end{table}

\begin{table}
  \centering \scriptsize
  \setlength\tabcolsep{3.3pt}
  \begin{tabular}{ccccccccccccc}
    \toprule
    \multirow{2}{*}{\Th{Group}} & \multirow{2}{*}{\Th{Efn}} & \multirow{2}{*}{\Th{Mean}} & \multirow{2}{*}{\Th{Gain}} & \Th{I2T} & \Th{T2I} & \Th{CR} & \Th{SCR} & \Th{TMIR}\\
    ~ & ~ & ~ & ~& R@1 &R@1 &Macro-F &Macro-F &R@10\\
    \midrule
    Random & ~ & 69.99 & ~ & 72.30 & 69.00 & 86.21  & 85.40  & 37.06  \\ 
   Hardest & ~ & 71.89  & \gain{+1.90} & 74.70 & 70.50 & 90.20 & 86.64 & 37.40 \\
    Hardest & \checkmark & 71.97 & \gain{+1.98} &74.30 & 71.00 & 90.10 & 87.15 & 37.28 \\
    \rowcolor{LightCyan}
     Semi-hard & \checkmark & \textbf{72.45} & \gain{+2.46} & \textbf{75.00} & \textbf{71.00} & \textbf{90.31} & \textbf{87.83} & \textbf{38.13} \\
    \bottomrule
  \end{tabular}
  \vspace{-8pt}
  \caption{Ablation study results comparing the \emph{Grouped Batch Sampling Strategy} across five downstream tasks. \Th{Group}: Strategy for grouping phase of GRIT. \Th{Efn}: Exclude False Negative in a grouping phase using index.} 
  \label{tab:Ablation2}
\end{table}

\section{Conclusion}
We introduced \textit{Synchronized attentional Masking} for enhanced masked modeling in fashion-centric VLMs. Leveraging cross-attention features of a momentum model, our method tailors the random mask into a targeted mask for synchronously co-occurring segments in image-text pairs in MLM and MIM objectives. This approach effectively resolves misaligned image-text input issues and improving fine-grained cross-modal representation. Additionally, we addressed data scarcity and distribution challenges in fashion datasets, refining grouped batch sampling with semi-hard negatives for ITM and ITC losses.
The experimental results showed our methods outperformed established benchmarks in multiple downstream tasks.

\section*{{Acknowledgment}}
\vspace*{-4pt}
This work was supported by Institute of Information \& communications Technology Planning \& Evaluation (IITP) under the metaverse support program to nurture the best talents (IITP-2024-RS-2023-00254529) grant funded by the Korea government (MSIT).

{\small
\bibliographystyle{ieee_fullname}
\bibliography{egbib}
}

\clearpage

\title{Supplementary material for \\ ``SyncMask: Synchronized Attentional Masking for Fashion-centric Vision-Language Pretraining''}

\maketitle
\thispagestyle{empty}

 \setcounter{page}{1}


\appendix

\renewcommand{\theequation}{A\arabic{equation}}
\renewcommand{\thetable}{A\arabic{table}}
\renewcommand{\thefigure}{A\arabic{figure}}

\section{Objectives for Vision-Language Pretraining}
In the realm of Vision-Language Pretraining (VLP), a variety of objectives are employed to effectively integrate vision-language features. This appendix delves into two such objectives, Image-Text Contrastive Learning (ITC) and Image-Text Matching (ITM), which were not extensively covered in the main text. For consistency, we use the same notation as employed in the main text.


\paragraph{Image-Text Contrastive Learning}
At the beginning of the multi-modal encoder, ITC aligns the shared latent space of the textual encoder and visual encoder. It employs a similarity function, denoted as $\sigma = g_v(z_v)^\top g_t(z_t)$ where $z_v$ is the \cls embedding of the vision encoder features $\bm{f}^{I}_{\theta}(V)$ and $z_t$ is derived from the text encoder features $\bm{f}^{T}_{\theta}(T)$. Then $g_v(\cdot)$ and $g_t(\cdot)$ mean linear projections which are mapping the \cls embeddings into normalized lower dimensional representations. We adopt the momentum contrastive learning~\cite{MoCo, ALBEF, FashionSAP} and leverage two queues that store $U$ recent uni-modal features obtained from the momentum model, denoted as $g'_v(z'_v)$ and $g'_t(z'_t)$ for their normalized features. Using this features, we formulate similarity as:
\begin{equation}
\label{itc_v2t}
\sigma(V,T) = g_v(z_v)^\top g'_t(z'_t) 
\end{equation} 
\vspace{-10pt}
\begin{equation}
\label{itc_t2v}
\sigma(T, V) = g_t(z_t)^\top g'_v(z'_v)
\end{equation}

For every matching set of text and image, the corresponding similarity measures are expressed as:
\begin{equation}
h^{v2t}_{k}(V) = \frac{\exp(\sigma(V, T_k)/\tau)}{\sum_{u=1}^{U} \exp(\sigma(V, T_u)/\tau)},
\end{equation}
\begin{equation}
h^{t2v}_{k}(T) = \frac{\exp(\sigma(T, V_k)/\tau)}{\sum_{u=1}^{U} \exp(\sigma(T, V_u)/\tau)}
\end{equation}

where $\tau$ denotes a learnable scaling factor. We define the ground-truth similarity using one-hot label as $\bm{y}^{v2t}(V)$ and $\bm{y}^{t2v}(T)$. Thus, ITC is constructed based on the cross-entropy, $\Th{CE}(\cdot, \cdot)$, between $\bm{h}$ and $\bm{y}$:
\begin{multline}
\mathcal{L}_{\Th{ITC}} = \frac{1}{2}\mathbb{E}_{(V,T) \sim D} [
\Th{CE}(\bm{y}^{v2t}(V),\bm{h}^{v2t}(V)) \\
+ \Th{CE}(\bm{y}^{t2v}(T),\bm{h}^{t2v}(T))]
\end{multline}

\paragraph{Image-Text Matching}

For the ITM loss, the model categorizes image-text pairs as either corresponding (positive) or non-corresponding (negative) by utilizing a shared representation derived from the output embedding of the $\cls$ token from the multi-modal encoder ${f^M_{\theta}({f^T_{\theta}(T)},{f^I_{\theta}(V)})}$. This vector is fed through a fully connected (FC) layer and softmax for binary classification,  producing the prediction probability $h^{itm}$. Similar to previous works~\cite{ALBEF, grit-vlp}, we exploit hard negatives in the ITM task, identifying pairs that exhibit akin meanings but vary in intricate specifics, employing in-batch contrastive similarity based on ITC. The ITM objective is expressed as:
\begin{equation}
   \mathcal{L}_\Th{ITM} = \mathbb{E}_{(V,T) \sim D} [\Th{CE}( \bm{y}^{itm}, \bm{h}^{itm}(V,T)], 
\end{equation}

\section{Semi-hard Negatives for Grouped Batch}
\paragraph{Grouped Mini-batch Sampling} GRIT-VLP~\cite{grit-vlp} employs an adaptive sampling strategy to gather similar samples within mini-batches, enhancing the effectiveness of mining hard negatives for both ITC and ITM. GRouped mIni-baTch sampling (GRIT) strategy consists of four phases: 1) Collecting, where \cls features $z_v$ and $z_t$ are stored in two queues, which are larger than a mini-batch size; 2) Example-level shuffling, which ensures randomness among the samples; 3) Grouping, where similar examples are grouped based on similarity scores; and 4) Mini-batch-level shuffling, which shuffles mini-batches for improving the model's ability to generalize. We tailor the 3) \textit{Grouping} phase to suit the characteristics of fashion data.

\begin{figure*}[t]
\begin{center}
	\fig[0.95]{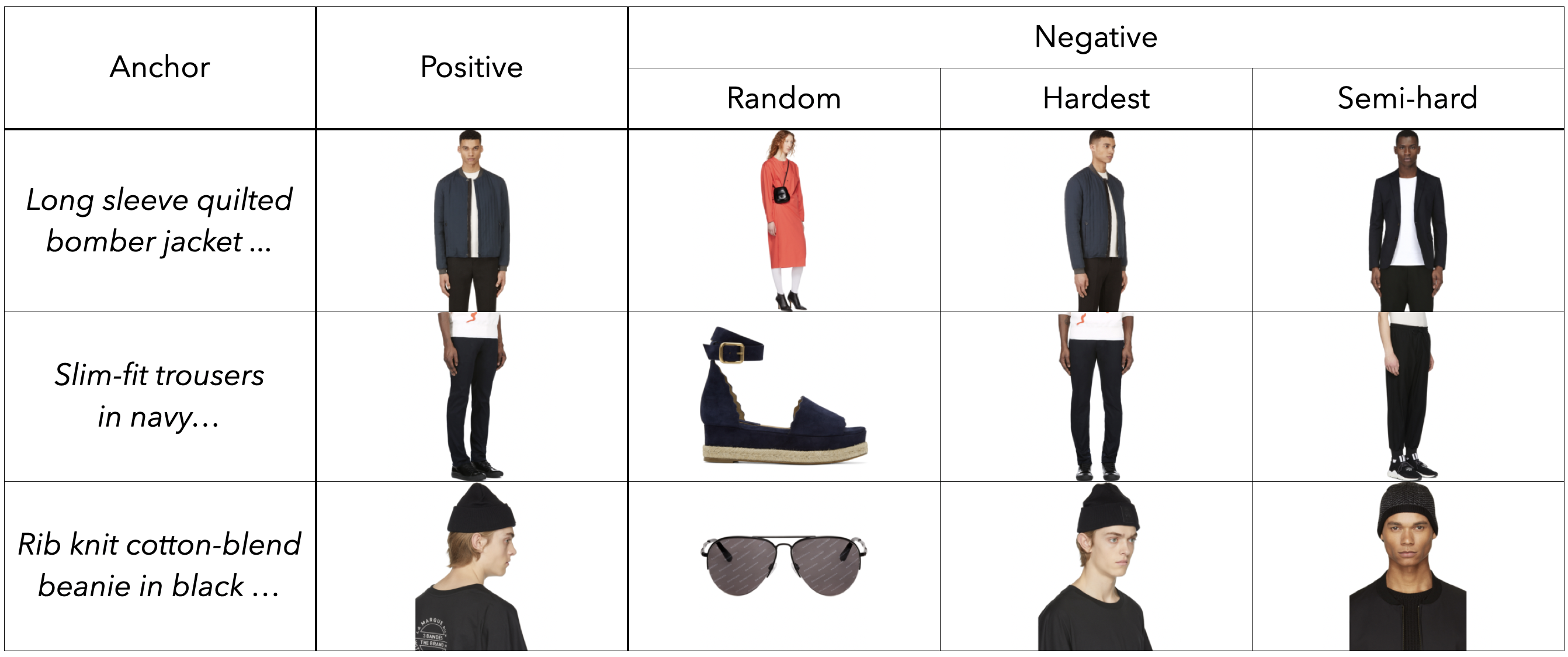}
\end{center}
\vspace{-8pt}
\caption{Illustration of the differences in negative samples selected within a mini-batch using various \textit{Grouping} methods: Random, Hardest and Semi-hard (Ours). }
\label{fig:semi_example}
\end{figure*}

In the Grouping phase, similarity scores for image-text pairs within sub-queues are calculated using methods similar to \autoref{itc_v2t} and \autoref{itc_t2v}. Each sub-queue has a size of 
$S$, smaller than the queue size of Collecting phase but larger than the mini-batch size. These scores denoted as $q^{v2t}(V) \in \mathbb{R}^S$ and $q^{t2v}(T) \in \mathbb{R}^S$, where $V$ and $T$ mean each image-text pair, are used to find similar examples. The process starts by randomly selecting an initial pair from the sub-queue. The algorithm then iteratively finds and stores the index of the example with the highest similarity score, one by one, in the index queue $I$. Instead of relying on a one-way similarity metric, the algorithm alternates between using image-to-text and text-to-image similarity scores. Thus, half of the pairs are selected based on their image-to-text similarity, while the remaining half are chosen based on their text-to-image similarity. By alternating between the two directions, the method ensures a balanced grouping, effectively capturing the relational nuances between images and texts in both directions. 

For the $(i+1)^{th}$ iteration, when considering a specific pair ($V_j, T_j$) with the index $j$, the selection of $I_{i+1}$ is determined as follows:
\begin{equation}
      I_{i+1} =
    \begin{cases}
       \mathtt{argmax}_{k \notin I} \bm{q}^{t2v}_{k}(T_j) & \text{if   } I_{i} \text{  is chosen with   }  \bm{q}^{v2t}\\
     \mathtt{argmax}_{k \notin I} \bm{q}^{v2t}_{k}(V_j) & \text{if   } I_{i} \text{  is chosen with   }  \bm{q}^{t2v}\\
    \end{cases}     
\label{eq:grouping}
\end{equation}
\paragraph{Refined Grouping for Semi-hard Negatives}
In the context of ITM and ITC tasks, the model is trained using both positive and negative images relative to a text anchor within a mini-batch. As illustrated in \autoref{fig:semi_example}, conventional VLP methods have relied on randomly selected negative images for training. This approach enables the model to identify negatives without necessarily learning fine-grained details. In contrast, the GRIT~\cite{grit-vlp} strategy constructs batches with hard negatives in a generic domain, facilitating more efficient learning by focusing on fine-grained information. 

However, in the fashion datasets, multiple images of the same item, differing only in angles, are often paired with a single text. Consequently, forming batches with the most similar samples can lead to the issue of false negatives. To address this, we have adjusted the grouping phase to include semi-hard negatives, which share similar features but are not identical. This adjustment ensures a more nuanced and effective training process, particularly suited to the unique challenges presented in the fashion domain. In our proposed modification in the Grouping phase, the algorithm shifts its focus from selecting the highest similarity score sample to choosing the $s^{th}$ most similar sample, where $s$ is a hyperparameter. We pre-train on the FashionGen~\cite{fashiongen} dataset, featuring three angle-specific images and one full-body image per text. Thus, observing reduced false negatives when $s$ is set to 3, we adopted this value for $s$. This method can be expressed as:
\begin{equation}
      I_{i+1} =
    \begin{cases}
       \mathtt{argsort}_{k \notin I} (\bm{q}^{t2v}_{k}(T_j))_s & \text{if   } I_{i} \text{  is chosen with   }  \bm{q}^{v2t}\\
     \mathtt{argsort}_{k \notin I} (\bm{q}^{v2t}_{k}(V_j))_s & \text{if   } I_{i} \text{  is chosen with   }  \bm{q}^{t2v}\\
    \end{cases}     
\label{eq:grouping}
\end{equation}

In this equation, the $\mathtt{argsort}$ function sorts the elements $\bm{q}^{t2v}_{k}(T_j)$ or $\bm{q}^{v2t}_{k}(V_j)$ in descending order, excluding indices already present in the index queue $I$, and $I_{i+1}$ is then assigned the $s^{th}$ index from this sorted list. This change aims to accumulate semi-hard negatives within the same mini-batch, thereby minimizing the false negatives in both ITC and ITM. To demonstrate the effectiveness of our method, we set all hyperparameters, including queue and sub-queue sizes, identical to those in GRIT-VLP~\cite{grit-vlp}, except for $s$.

\end{document}